# THE NEW HYBRID COAW METHOD FOR SOLVING MULTI-OBJECTIVE PROBLEMS

Zeinab Borhanifar and Elham Shadkam[*]

Department of Industrial Engineering, Faculty of Eng.; Khayyam University, Mashhad, Iran

## ABSTRACT

*In this article using Cuckoo Optimization Algorithm and simple additive weighting method the hybrid COAW algorithm is presented to solve multi-objective problems. Cuckoo algorithm is an efficient and structured method for solving nonlinear continuous problems. The created Pareto frontiers of the COAW proposed algorithm are exact and have good dispersion. This method has a high speed in finding the Pareto frontiers and identifies the beginning and end points of Pareto frontiers properly. In order to validation the proposed algorithm, several experimental problems were analyzed. The results of which indicate the proper effectiveness of COAW algorithm for solving multi-objective problems.*

## KEYWORDS

*Cuckoo Optimization Algorithm (COA), simple additive weighting (SAW), Pareto frontier, Multi-objective optimization problem (MOP).*

## 1. INTRODUCTION

There are many methods for solving nonlinear constrained programming problems such as Newton, Genetic algorithm, the algorithm of birds and so on. In this paper using the emerging Cuckoo Optimization Algorithm and simple additive weighting a method to solve multi-objective problems is presented.

In single-objective optimization, it is assumed that the decision makers communicate only with one goal like: profit maximization, cost minimization, waste minimization, share minimization and so on. But in the real world it is not possible to consider single goals and usually more than one goal are examined. For example, in the control of the projects if only the time factor is considered, other objectives such as cost and quality are ignored and the results are not reliable. So it is necessary to use multi-objective optimization problems.

Ehrgott and Gandibleux presented a detailed approximation method regarding the problems related to combinatorial multi-objective optimization [1]. Klein and Hannan for multiple objective integer linear programming problems (MOILP) presented and algorithm in which some additional restrictions is used to remove the known dominant solutions [2]. Sylva and Crema offered a method to find the set of dominant vectors in multiple objective integer linear programming problems [3]. Arakawa et al. used combined general data envelopment analysis and Genetic Algorithm to produce efficient frontier in multi-objective optimization problems [4].

Deb analyzed the solution of multi-objective problems by evolutionary algorithms [5]. Reyes-seerra and Coello Coello analyzed the solution of multi-objective problems by particle swarm [6]. Cooper et al. have worked on the solution of multi-objective problems by the DEA and presenting an application [7]. Pham and Ghanbarzadeh solved multi-objective problems by bee algorithm [8]. Nebro et al. analyzed a new method based on particle swarm algorithm for solving multi-objective optimization problems [9]. Gorjestani et al. proposed a COA multi objective algorithm using DEA method [10].

For multi-objective optimization problems usually it is not possible to obtain the optimal solution that simultaneously optimizes all the targets in question. Therefore we should try to find good solutions rather than the optimal ones known as Pareto frontier. Given that so far the Simple Additive Weighting method is not used in meta-heuristic, especially cuckoo algorithms, this paper presents a combined method.

The first section introduces Cuckoo optimization algorithm, then in the second section Simple Additive Weighting (SAW) method is discussed as a combined method for solving multi-objective described. Finally, the fourth section provides the proposed implemented approach, numerical results and a comparison which is made with other methods.

## 2. CUCKOO OPTIMIZATION ALGORITHM

Cuckoo optimization algorithm was developed by Xin-She Yang and Suash Deb in 2009. Thence Cuckoo optimization algorithm was presented by Ramin Rajabioun in 2011 [11]. Cuckoo algorithm flowchart is as figure 1. This algorithm applied in several researches such as production planning problem [12], portfolio selection problem [13], evaluation of organization efficiency [14], evaluation of COA [15] and so on. For more information about the algorithm refer to [11].

## 3. SIMPLE ADDITIVE WEIGHTING METHOD

SAW is one of the most practical methods designed for decision-making with multiple criteria presented by Hong and Eun in 1981. In this method which is also known as weighted linear combination after scaling the decision matrix by weighted coefficients of criteria, the free scale weighted decision matrix id obtained and according to this scale the score of each option is selected. The most important feature of this method is the simple application because of its mathematical logic.

Assuming the multiple target model (1) and defining the parameters $w_1$ and $w_2$ which are the weight of the objective functions and defined based on the importance of the functions by the decision maker, the model can be converted to single-objective models (2):

$$Max\ F(x)=(f_1(x), f_2(x), \ldots, f_3(x))$$
$$s.t.\ g_i(x) \leq b_i \tag{1}$$
$$x_i \geq 0$$
$$Max\ F(x)= w_1 f_1 + w_2 f_2 + \cdots + w_n f_n$$
$$w_1 + w_2 + \cdots + w_n = 1 \tag{2}$$

In these models $f_1(x) \ldots f_k(x)$ are objective functions. $w_i$ is the weight defined by the importance of the decision maker.

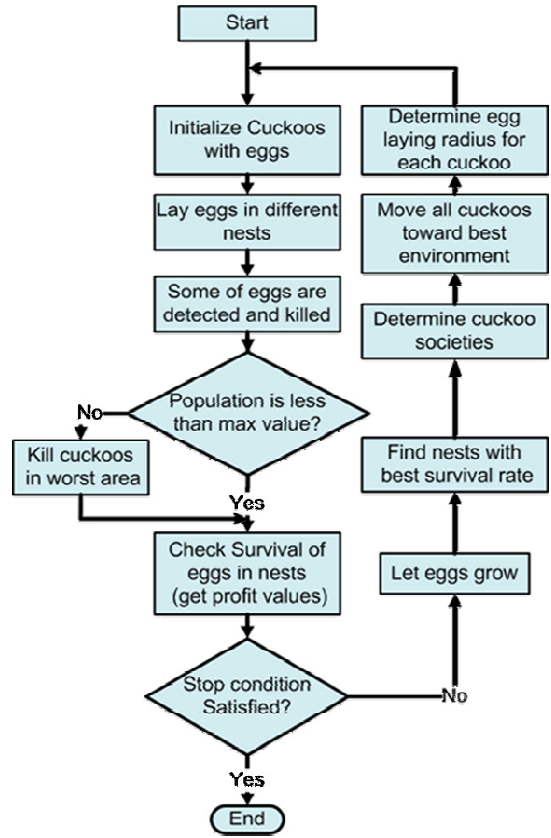

Figure 1: The Cuckoo optimization algorithm flowchart

## 4. PRESENTATION OF HYBRID COAW ALGORITHM

In this section we present the method COAW which is proposed in this paper. The steps of this algorithm are as follows. Also the flowchart of COAW algorithm is as figure 2.

Step1 Different random $w_1$ and $w_2$ are generated subject to the summation of these two values equals to one.

Step 2 The present locations of Cuckoos are determined randomly

Step 3 A number of eggs are allocated to each Cuckoo

Step 4 The laying radius of each Cuckoo is determined

Step 5 The Cuckoos hatch in the nests of the hosts that are within their laying radius

Step 6 Eggs that are detected by the host birds are destroyed

Step 7 The eggs of the identified cuckoos are nurtured

Step 8 The habitats of the new cuckoos are evaluated by SAW method and determined weights

Step 9 the maximum number of cuckoos living at each location are determined and the ones in wrong areas are destroyed

Step 10 The cuckoos are clustered by K-means and the best cluster of cuckoos is determined as the residence

Step 11 The new population of cuckoos moves toward the target location

Step 12 Stop condition is established otherwise go to step 2Step 13 the value of $f_1, f_2$ are determined for the best solutions and the Pareto frontier is gained based on $f_1, f_2$

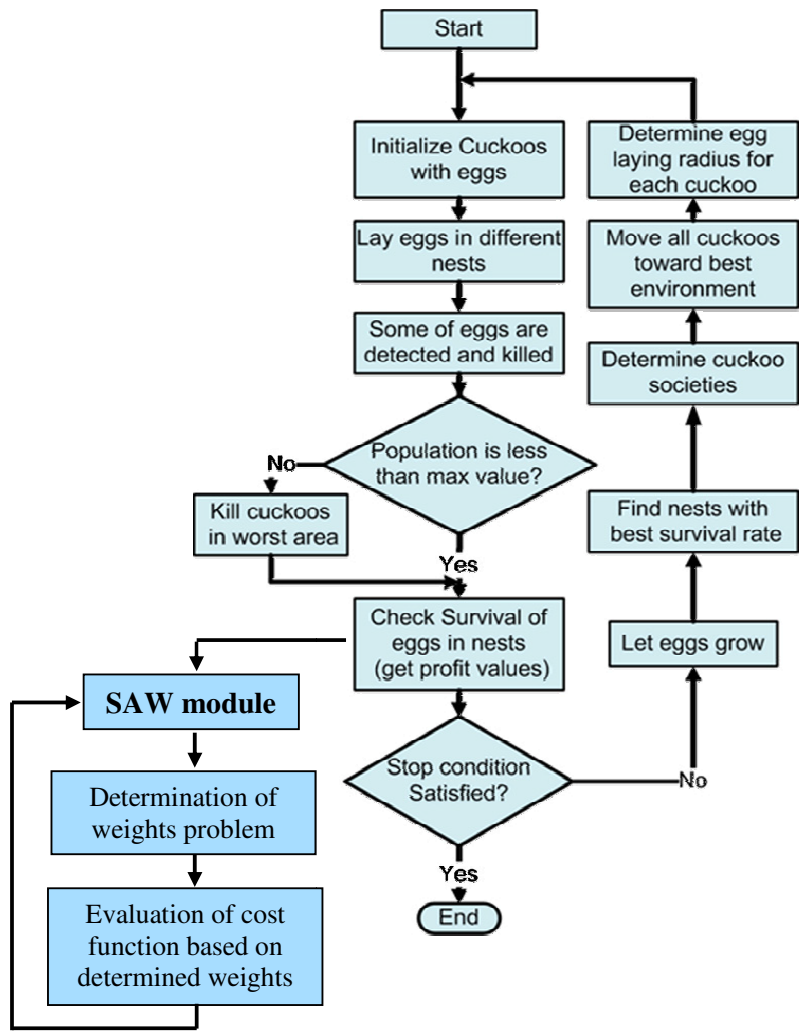

Figure 2. The flowchart of COAW algorithm

## 5. IMPLEMENTATION OF COAW ALGORITHM ON SOME TEST PROBLEMS

In this section in order to validation the COAW algorithm some test problems are analyzed. Test problems are presented in Table 1.

Table 1. Test problems

| Number of problem | Objectives | Constraints |
|---|---|---|
| 1 | $Min \, f_1(x) = x_1$ <br> $Min \, f_2(x) = x_2$ | $(x_1 - 2)^2 + (x_2 - 2)^2 - 4 \leq 0$ <br> $x_1, x_2 \geq 0$ |
| 2 | $Min \, f_1(x) = 2x_1 - x_2$ <br> $Min \, f_2(x) = -x_2$ | $(x_1 - 1)^3 + x_2 \leq 0$ <br> $x_1, x_2 \geq 0$ |
| 3 | $Min \, f_1(x) = x_1$ <br> $Min \, f_2(x) = x_2$ | $x_1{}^3 - 3x_1 - x_2 \leq 0$ <br> $x_1 \geq -1, x_2 \leq 2$ |

Given that determining input parameters is one of the effective problems in meta-heuristic algorithms, so the parameters of the algorithm are presented as follows: the number of initial population=5, minimum number of eggs for each cuckoo= 2, maximum number of eggs for each cuckoo =4, maximum iterations of the Cuckoo Algorithm=50, number of clusters that we want to make=1, Lambda variable in COA paper=5, accuracy in answer is needed=-3.75, maximum number of cuckoos that can live at the same time=10, Control parameter of egg laying=5, cuckooPopVariance = 1e-13.

## 6. THE SOLUTION OF TEST PROBLEMS

In this section the experimental problems of the previous section are solved by the proposed algorithm and the results are compared and examined with the same algorithm.

### 6.1. The First Problem

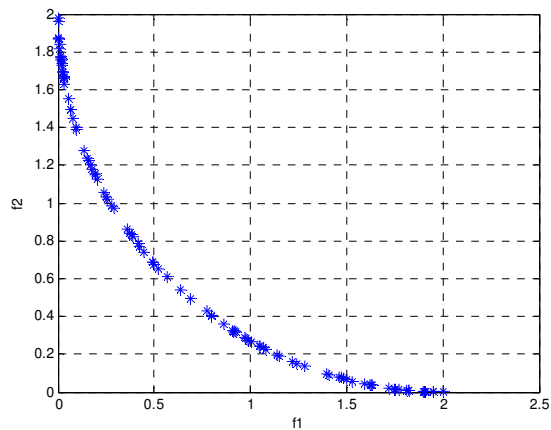

Figure 3. Pareto frontiers created by COAW algorithm for first problem

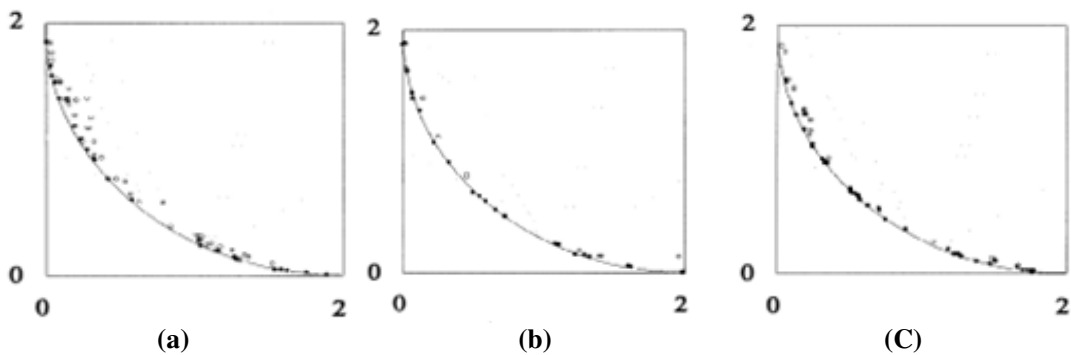

Figure 4. Pareto frontiers created by: (a) Ranking method (b) DEA method (c) GDEA Method for first problem

## 6.2. The Second Problem

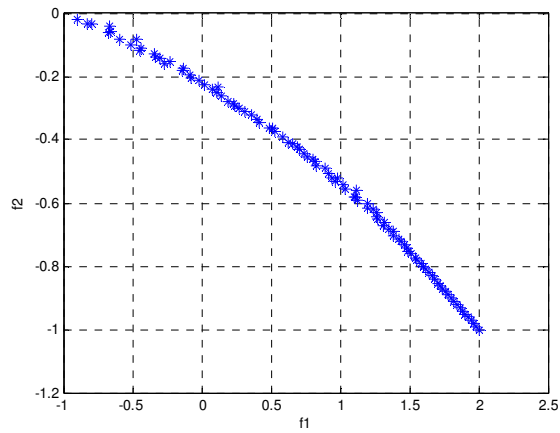

Figure 5. Pareto frontiers created by COAW for the first problem

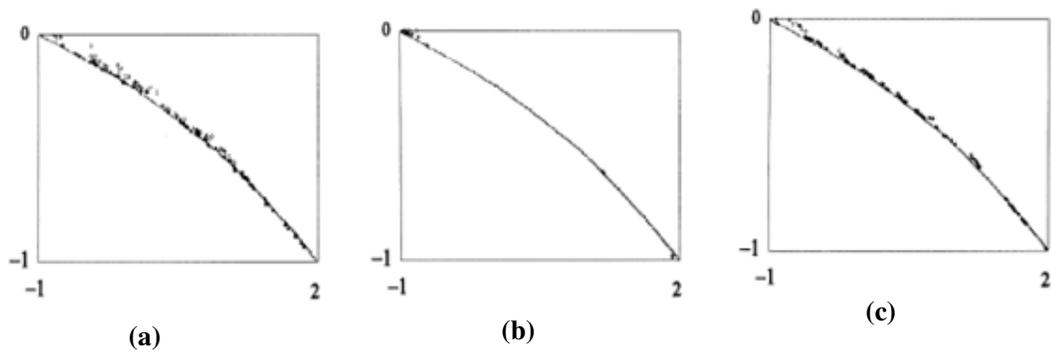

(a)          (b)          (c)

Figure 6. Pareto frontiers created by: (a) Ranking method (b) DEA method (c) GDEA Method for second problem

## 6.3. The Third Problem

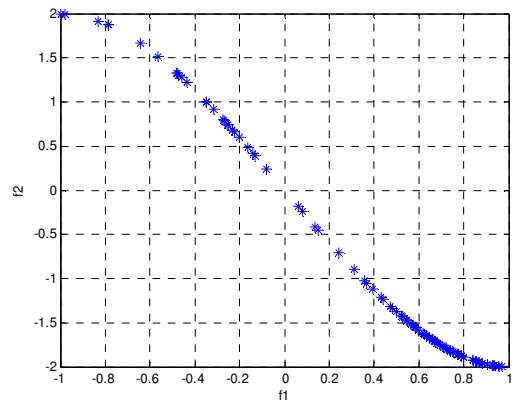

Figure 7. Pareto frontiers established by COAW for third problem

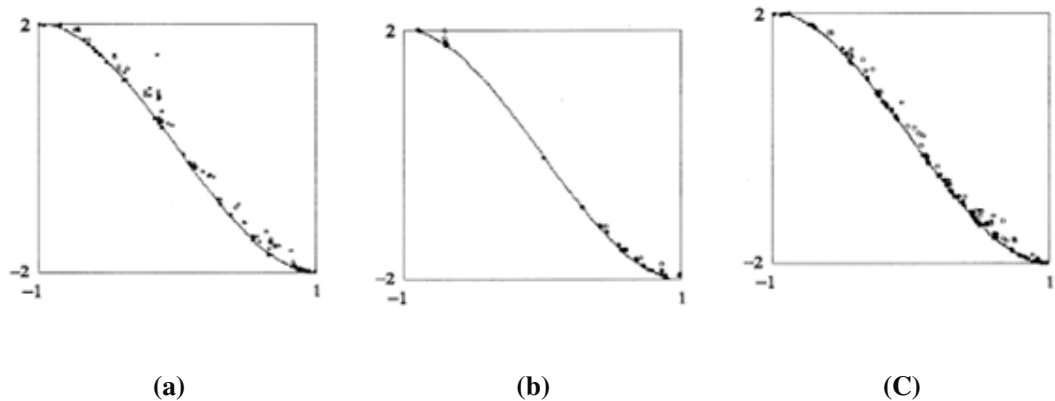

|        |        |        |
|:------:|:------:|:------:|
| **(a)** | **(b)** | **(C)** |

Figure 8. Pareto frontiers created by: (a) Ranking method (b) DEA method (c) GDEA Method for third problem

After the implementation of the proposed approach on test problems the Pareto frontiers are obtained according to figures 3, 5 and 7 in order to compare the COAW method with other methods, ranking method, DEA method and GDEA method are implemented on problems. The results are show as figures 4, 5 and 8.

As figures indicate the created Pareto frontiers of the COAW proposed algorithm are exact and have good dispersion. This method has a high speed in finding the Pareto frontiers and identifies the beginning and end points of Pareto frontiers properly. The COAW algorithm not only solves the problems with lower initial population 5 but also it presents better and more exact answers in fewer repetitions than similar methods.

## 7. CONCLUSION

In this paper the hybrid COAW algorithm was presented to solve multi-objective problems. The hybrid approach includes Cuckoo Algorithm and Simple Additive Weighting method. The algorithm was analyzed for a number of experimental problems and compared with several similar methods. The results indicate the accuracy in finding Pareto frontiers. Also the Pareto frontier is better than similar methods and as a result COAW proposed method is reliable, fast and simple to solve multi-objective optimization problems.